\documentclass[runningheads]{llncs}
\usepackage[dvipdfmx]{graphicx}
\usepackage{color}
\usepackage{amsmath}
\newcommand{\red}[1]{\textcolor{red}{#1}}

\usepackage{array}
\usepackage{amssymb}

\begin{document}
\title{Font Style Interpolation with Diffusion Models}

\author{Tetta Kondo\and
Shumpei Takezaki\and
Daichi Haraguchi\orcidID{0000-0002-3109-9053}\and
Seiichi Uchida\orcidID{0000-0001-8592-7566}}
\authorrunning{T. Kondo et al.}

\institute{Kyushu University, Fukuoka, Japan\\
\email{\{shumpei. takezaki, seiichi.uchida\}@human.ait.kyushu-u.ac.jp}}
\maketitle              
\begin{abstract}
Fonts have huge variations in their styles and give readers different impressions. Therefore, generating new fonts is worthy of giving new impressions to readers. In this paper, we employ diffusion models to generate new font styles by interpolating a pair of reference fonts with different styles. More specifically, we propose three different interpolation approaches, image-blending, condition-blending, and noise-blending, with the diffusion models. We perform qualitative and quantitative experimental analyses to understand the style generation ability of the three approaches. According to experimental results, three proposed approaches can generate not only expected font styles but also somewhat serendipitous font styles. We also compare the approaches with a state-of-the-art style-conditional Latin-font generative network model to confirm the validity of using the diffusion models for the style interpolation task. 
\end{abstract}
%
%
%
\section{Introduction}
Fonts have huge variations in their styles. For example, {\tt MyFonts.com} says they have over 270,000 fonts in their collection. Even today, font design experts generate new fonts with new styles. At {\tt MyFonts.com}, each font is related to multiple tags; for example, a famous font {\tt Helvetica} is related to {\it modern}, {\it modest}, {\it neutral}, and so on. This means different fonts give readers different impressions; therefore, generating new fonts is worthy of giving new impressions to them.\par
Many machine-learning technologies have been developed for automatically generating fonts in various styles. Generative Adversarial Networks (GANs) and Variational Auto-Encoders (VAEs) are conventional neural network models for font generation~\cite{srivatsan2019deep,wang2020attribute2font,xie2021dg}. These models can generate diverse styles using random vectors as their inputs or conditions. One-shot (or few-shot) font generation uses one (or several) character examples (say, `A') in a certain style to generate the remaining characters (`B'-`Z') in the same style~\cite{srivatsan2019deep,xie2021dg}. These technologies are often based on disentanglement, which decomposes each character image into style and character-class information.
\par 

Interpolation is a classical way to generate new font styles from a pair of reference fonts, $\mathbf{r}_1$ and $\mathbf{r}_2$, with different styles.
One of the pioneering attempts for interpolation-based font generation is Campbell and Kauts~\cite{Campbell2014}, which forms a manifold of font style variations and performs style interpolation operations on it. 
Interpolation possesses the advantage of intuitively controlling the generated style. We can roughly imagine their interpolated versions from the given pair of fonts. In other words, we can generate fonts in the ``expected'' style relatively easily by choosing a pair of fonts similar to the desired style. Of course, by aiming for serendipity, it is possible to interpolate random font pairs to discover novel styles.\par 
\begin{figure}[t] 
\includegraphics[width=\textwidth]{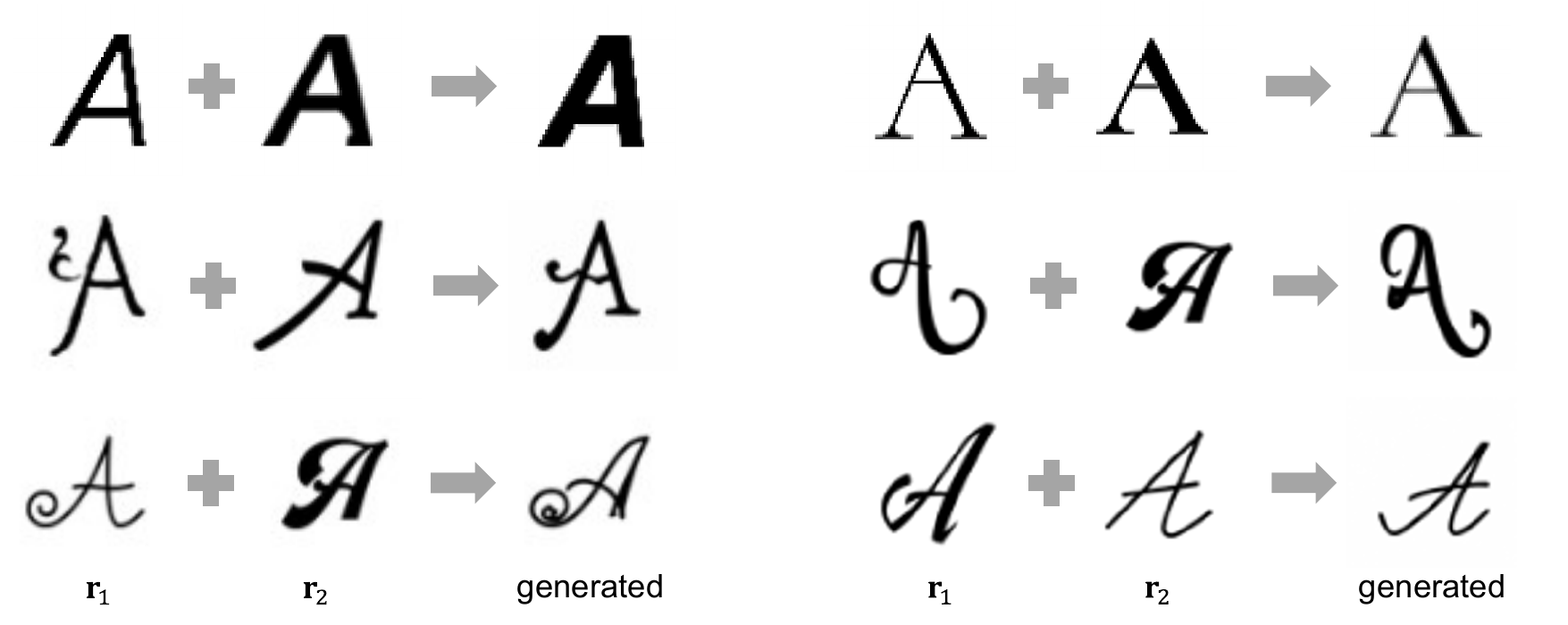}
\caption{Examples of font style interpolation by our approach, called noise-blending. 
The reference images $\mathbf{r}_1, \mathbf{r}_2$ in the top row: Google Fonts. Other rows: MyFonts.} \label{fig:interpolation-example}
\end{figure} 

Fig.~\ref{fig:interpolation-example} shows examples of fonts generated by an interpolation method proposed in this paper. If we choose a similar font pair, the interpolated results show medium styles between the pair. The results show rather unexpected styles if we choose a different font pair. These results suggest that interpolation is a reasonable way not only to generate expected styles but also to generate somewhat serendipitous styles.\par

\begin{figure}[t] 
\includegraphics[width=\textwidth]{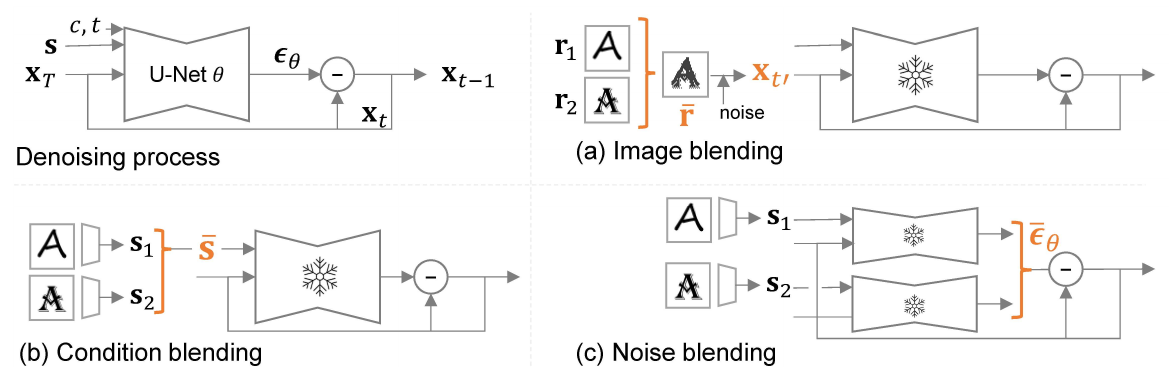}\\[-2mm]
\caption{Overview of the denoising process and our three approaches for font style interpolation: (a)~Image blending. (b)~Condition blending. (c)~Noise blending. For simplicity, several operations (constant multiplications and addition of stochastic perturbation) are omitted in the denoising process. In (a)-(c), unimportant conditions $t, c$ are also omitted} \label{fig:overview}
\end{figure}

Diffusion models are becoming more popular for image generation than GANs and VAEs. Although diffusion models are also based on neural networks, they take a very different strategy for image generation. Specifically, diffusion models use a stochastic and iterative denoising process to generate realistic images from random noise images. As shown in Fig.~\ref{fig:overview}~Denoising process, the denoising process is realized by a U-Net trained to estimate the noise component of its input image. Subtracting this estimated noise from the input image yields a less noisy image, which becomes the input for the next iteration.
\par

The purpose of this paper is to tackle the font style interpolation task 
with diffusion models. We can expect several merits of diffusion models for the task: 
\begin{itemize}
    \item Diffusion models generate realistic and decorative font images. In fact, diffusion models have already been applied to font (or character image) generation tasks~\cite{Tanveer_2023_ICCV} and proved their ability to realize not just realistic but also very decorative font images. 
    \item Except for the models in latent spaces, such as StableDiffusion~\cite{rombach2022high}, diffusion models perform operations in the original image domain (i.e., $W\times H$-dimensional space for $W\times H$-pixel images). Therefore, it is possible to incorporate some pixel-level controls. 
    \item Diffusion models are very flexible and versatile. For example, they can be character class-conditional to generate character images of a specific class $c$. Moreover, they allow operations on the estimated noises; for example, if we merge a noise image and its 180-degree rotated version, the model for character images generates ambigrams~\cite{shirakawa2023ambigram}, which are character images with dual readability.\par
\end{itemize}

To fully utilize the flexibility and versatility of the diffusion models, we propose and compare three different style interpolation approaches,  image-blending, style feature-blending, and noise-blending, with diffusion models. Fig.~\ref{fig:overview} is an overview of these approaches. 
\begin{itemize}
\item {\em Image-blending approach} performs the interpolation operation in the image domain. It first prepares a blended image of two character images in different styles and then generates a realistic image from it by employing the idea of SDEdit~\cite{meng2021sdedit}.
\item {\em Condition-blending approach} interpolates style conditions of a conditional diffusion model. The neural network for the denoising process is trained under a font style condition; in other words, the diffusion model is trained to generate font images specified by a style condition. The style condition is represented as a real-valued vector (instead of a one-hot vector). This approach tries to generate intermediate styles with interpolated style condition vectors. 
\item {\em Noise-blending approach} interpolates (still noisy) images under their denoising process. Recent trials, such as \cite{shirakawa2023ambigram}, found that operations on the intermediate images in the denoising process affect the final results of the process. In this approach, we expect blending the noisy images will yield a result with blended styles.
\end{itemize}

We perform qualitative and quantitative experimental analyses to understand the style generation ability of the three approaches and the effect of paired font images. We compare the performance of the proposed approaches to the state-of-the-art style-conditional Latin-font generative network model, FANnet~\cite{Roy_2020_CVPR}, to prove how style interpolation with diffusion models is useful. 
\par

\section{Related Work}
\subsection{Font Style Features} 
As pointed out from 1920s~\cite{davis1933determinants,poffenberger1923study}, font styles play crucial roles for transmitting impressions and enhancing legibility. Recently, deep neural networks enable us to deal with the styles as font style feature vectors, which are utilized in various font-related tasks. DeepFont~\cite{wang2015deepfont} is an early trial to classify font images into their font class (such as Helvetica)  by a neural network, in which font style features are implicitly extracted for the classification. More recent neural network models extract the style features more explicitly and thus realize style occurrence analysis~\cite{yasukochi2023analyzing} and style recommendation~\cite{choi2019assist,kulahcioglu2020fonts}. Font generation is the most widely tackled task with font style features. A typical task setup is one or few-shot font generation, where the font style features of the one or several example character images (say, `A' and `B') are first extracted and then used for generating the remaining character images (`C' to `Z'). The disentanglement technique is often introduced to separate the font style and character features. Srivatsan et al.~\cite{srivatsan2019deep,srivatsan2021scalable} proposed a VAE-based model to disentangle font styles from font images. AGIS-Net~\cite{gao2019artistic} is a GAN-based approach to generate more artistic fonts, including character color and texture. AGIS-Net disentangles font style features and character features by two separated encoders: a style encoder and a content encoder. Note that few-shot font generation is very important for character-rich languages, such as Chinese and Korean~\cite{Kong_2022_CVPR,Liu_2022_CVPR,Wang_2023_CVPR}, because designing thousands of character images is necessary for each font set.\par

FANnet~\cite{Roy_2020_CVPR} is a state-of-the-art one-shot font generation model, especially for Latin alphabets, despite its simple auto-encoder structure for explicitly extracting font style feature vectors. As we will see in Fig.~\ref{fig:style-feature-extraction}~(a), the style feature vector is fed into a class-conditional decoder to generate a character image of the specified class in the same style as the encoder input. This paper uses FANnet to extract style features $\mathbf{s}$ for interpolation approaches (excluding the image-blending approach). Additionally, we use FANnet as a comparison model in our experiments.

\subsection{Font Generation by Diffusion Models}
Diffusion models have already been applied to two types of font generation tasks. The first task is one or few-shot font generation. Diff-Font~\cite{he2022diff} and FontDiffuser~\cite{yang2024fontdiffuser} generate Chinese font images by diffusion model in the few-shot approach. The other is artistic typography (or font) generation~\cite{IluzVinker2023,Tanveer_2023_ICCV,wang2023anything}. They employ recent prompt-based output control. For example, DS-Fusion~\cite{Tanveer_2023_ICCV} generates the cat-like character image with the input prompt ``cat.'' Wang \textit{et al}. also generate artistic font consisting of ``pasta'' by the prompt ``pasta.'' These diffusion models for font generation show the great ability to generate various decorative font images.\par

Style transfer is a recent application of diffusion models and may be applicable to font image generation. Most of the style transfer tasks aim to transfer mainly the texture and color features (rather than shape features) of a given natural image to another image. StyleDiffusion~\cite{Wang_2023_ICCV} proposes a disentanglement scheme to decompose a reference natural object image into style and content features. StainDiff~\cite{staindiff} can transfer the stain styles of a histology image into other images. Diffusion models have also been used for text-guided style transfer in text-to-image synthesis. ControlStyle~\cite{controlstyle}, similar to other studies~\cite{ahn2023dreamstyler,Hamazaspyan_2023_CVPR,DiffStyler,Pan_2023_WACV,SGDiff,yang2023zero}, generates photo-realistic images by prompts, while reflecting the style of a reference image.\par

To the authors' best knowledge, no diffusion model has been applied to shape interpolation, especially font style interpolation. Moreover, as reviewed through the above recent style transfer models,
most diffusion models for image-to-image conversions focus more on textures and colors. Our idea of combining this ability with style interpolation (i.e., one of the shape conversion tasks) is expected to realize an intuitive brand-new font design framework.

\section{Three Approaches for Font Style Interpolation with Diffusion Models}
\subsection{Conditional Diffusion Model for Character Image Generation}  
All three approaches for style interpolation use the same pretrained diffusion model. Just as the standard diffusion model, our model generates a character image $\mathbf{x}_0$ from a noise image $\mathbf{x}_T\sim \mathcal{N}(\mathbf{0}, \mathbf{I})$ by iteratively generating denoised images $\mathbf{x}_{T-1},\ldots,\mathbf{x}_t,\ldots,\mathbf{x}_1, \mathbf{x}_0$. As illustrated in Fig.~\ref{fig:overview}~Denoising process, this denoising process is realized by a conditional U-Net architecture based on~\cite{ho2020denoising}. Specifically, the U-Net with the weight parameter set $\theta$ estimates the noise component $\boldsymbol{\epsilon}_\theta(\mathbf{x}_t \mid t, c, \mathbf{s})$ in the (noisy) image $\mathbf{x}_t$, where $c$ is the character class (e.g., `A' and `B') and $\mathbf{s}$ is the real-valued condition vector specifying a font style. Then, a bit less noisy image $\mathbf{x}_{t-1}$ is obtained by subtracting $\boldsymbol{\epsilon}_\theta(\mathbf{x}_t \mid t, \mathbf{c}, \mathbf{s})$ from $\mathbf{x}_t$. The subtraction operation is the same as the standard denoising equation~\cite{ho2020denoising}, and its detail is omitted here.\par

The style condition $\mathbf{s}$ is important for our font image generation task. As detailed in Section~\ref{sec:style-feature-extraction}, we use a convolutional neural network (CNN) model to obtain a style feature vector of each font. This style feature vector is used as the condition $\mathbf{s}$. When $\mathbf{s}$ is specified as the style condition, the U-Net $\theta$ is trained to generate the character images of the corresponding font style. Note that when a so-called ``null token'' \cite{ho2022classifier} is used as $\mathbf{s}$, the U-Net practically works in the unconditional generation mode. (The null token is trained along with the U-Net.) The character class condition $c$ is converted to a one-hot vector and then fed to the U-Net.\par

After the U-Net (i.e., the diffusion model) is sufficiently trained to generate images $\mathbf{x}_0$ of character class $c$ in style $\mathbf{s}$, all the weights $\theta$ are frozen in the following style interpolation process. In other words, no further fine-tuning step is introduced in individual interpolation approaches. This fact is practically beneficial; we can compare the style interpolation results from all three approaches just by using the common U-Net model. 

\subsection{Image-Blending Approach}  
As shown in Fig.~\ref{fig:overview}~(a), the image-blending approach uses a blended image of two reference character images, $\mathbf{r}_1$ and $\mathbf{r}_2$. 
Here, image-blending is performed by the pixel-wise {\tt OR} operation, that is, 
$$
\overline{\mathbf{r}}=\mathbf{r}_1\oplus\mathbf{r}_2.     
$$
Obviously, the blended image $\overline{\mathbf{r}}$ has an unnatural character shape, especially when two reference images have very different styles. However, if we use $\overline{\mathbf{r}}$ as an initial image and then generate a natural character image from it in the denoising process, we can expect that the generated image $\mathbf{x}_0$ reflects the styles of $\mathbf{r}_1$ and $\mathbf{r}_2$.\par

For generating $\mathbf{x}_0$ from $\overline{\mathbf{r}}$, we employ the idea of SDEdit~\cite{meng2021sdedit}. SDEdit does not start its denoising process from $t=T$ with the pure noise image $\mathbf{x}_T$. Instead, it starts the denoising process from $t=t'<T$ with a ``fake'' $\mathbf{x}_{t'}$. The term ``fake'' means 
that $\mathbf{x}_{t'}$ is not the actual image by the denoising process from $t=T$ to $t'$ but an artificial image. Here,  we use 
$\mathbf{x}_{t'}=\overline{\mathbf{r}}+\mathbf{z}$, where $\mathbf{z}$ is a noise vector so that $\mathbf{x}_{t'}$ mimics a (noisy) image at the denoising process $t=t'$. Since the U-Net is trained to generate natural character images finally, 
the denoising process from $t=t'$ to $0$ will result in a natural character image $\mathbf{x}_0$ that reflects the styles of $\mathbf{r}_1$ and $\mathbf{r}_2$. In the latter experiment, we set $t'=500$ and $T=1,000$. In the image-blending approach, no style condition is specified to utilize the original style of the blended image; as noted above, we use a null token as the style condition $\mathbf{s}$ that gives no effect.
\par

\subsection{Condition-Blending Approach}  
As shown in Fig.~\ref{fig:overview}~(b), the condition-blending approach uses a blended condition $\overline{\mathbf{s}}$ to specify the style.
Assume that we have the style feature vectors $\mathbf{s}_1$ and $\mathbf{s}_2$ of two reference character images, $\mathbf{r}_1$ and $\mathbf{r}_2$, respectively, by a CNN described in Section~\ref{sec:style-feature-extraction}. Then, the blended condition is given as  
$$
    \overline{\mathbf{s}}=\lambda\mathbf{s}_1+(1-\lambda)\mathbf{s}_2. 
$$
If the effect of the style condition  $\mathbf{s}$ is continuous (that is, if a small change of $\mathbf{s}$ appears as a small change of the generated image $\mathbf{x}_0$), we can expect that the diffusion model generates an interpolated image of  $\mathbf{r}_1$ and $\mathbf{r}_2$ by setting $\lambda\in (0,1)$ at a certain value.

\subsection{Noise-Blending Approach}  
As shown in Fig.~\ref{fig:overview}~(c), the noise-blending approach uses a blended noise $\overline{\boldsymbol{\epsilon}}_\theta$ for style interpolation.
Recent research, such as \cite{shirakawa2023ambigram}, proved that the estimated noise $\boldsymbol{\epsilon}_\theta$ can be blended to control output images.
We, therefore, blend two estimated noise images for two styles $\mathbf{s}_1$ and $\mathbf{s}_2$, namely, 
$$
    \overline{\boldsymbol{\epsilon}}_\theta=\lambda\boldsymbol{\epsilon}_\theta(\mathbf{x}_t \mid t, c, \mathbf{s}_1)+(1-\lambda)\boldsymbol{\epsilon}_\theta(\mathbf{x}_t \mid t, c, \mathbf{s}_2),
$$ 
where the two style conditions $\mathbf{s}_1$ and $\mathbf{s}_2$ are style feature vectors of 
two reference images, $\mathbf{r}_1$ and $\mathbf{r}_2$, respectively, similar to the condition-blending approach, and $\lambda\in (0,1)$. We have a denoised image $\mathbf{x}_{t-1}$ using this blended noise. By repeating this denoising process with $\overline{\boldsymbol{\epsilon}}_\theta$ from $t=T$ to $0$, the model generates $\mathbf{x}_0$ with an interpolated style of $\mathbf{r}_1$ and $\mathbf{r}_2$.

\begin{figure}[t] 
\includegraphics[width=\textwidth]{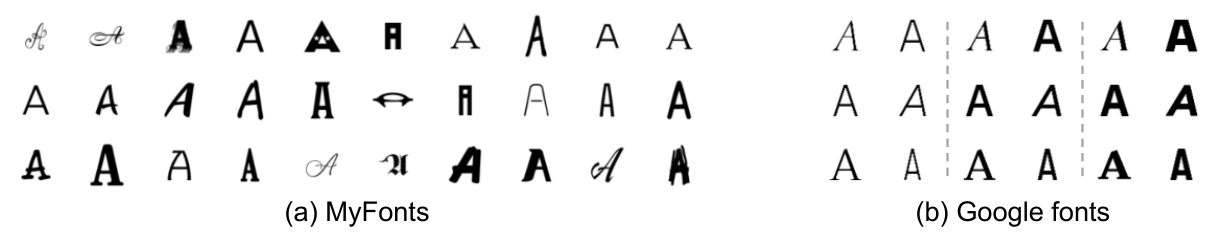}\\[-5mm]
\caption{Character images in various font styles.} \label{fig:style-examples}
\end{figure} 
\section{Experimental Results}
\subsection{Datasets\label{sec:datasets}}  
For the experimental evaluations, we use 26 capital Latin alphabet images (`A'--`Z') from MyFonts dataset~\cite{chen2019large} and GoogleFonts dataset~\footnote{\url{https://github.com/google/fonts}}.
Fig.~\ref{fig:style-examples} shows examples of character images from both datasets. As an image preprocessing, individual character images are resized to $64\times 64$ pixels.
\par

As shown in Fig.~\ref{fig:style-examples}~(a), the MyFonts dataset comprises various fonts, including standard fonts and very decorative fonts. We use this dataset for training diffusion models as well as FANnet and for performance evaluations. More specifically, we split 17,412 fonts in the dataset into 13,938 training fonts, 1,734 validation fonts, and 1,740 test fonts. The validation set is used to adjust the hyperparameter (learning rate) and terminate the training process for FANnet. The test set is used for quantitative and qualitative evaluations.\par

As shown in Fig.~\ref{fig:style-examples}~(b), the GoogleFonts dataset has less style diversity than the MyFonts dataset. We use this dataset as an additional test set for performance evaluation (i.e., not for training). This dataset has two benefits. First, each font is annotated with one of four style categories: Serif, Sans-serif, Handwriting, and Display. We, therefore, can evaluate the category-wise interpolation performance by this dataset. Second, the GoogleFonts dataset contains font families with different weights (i.e., stroke width). For example, it contains {\tt AlegreyaSans-Light}, {\tt AlegreyaSans-Medium}, and  {\tt AlegreyaSans-Bold}, which have different weights but belong to the same font family {\tt AlegreyaSans}. Fig.~\ref{fig:style-examples}~(b) shows light, medium, and bold versions of six fonts. The GoogleFonts dataset comprises 2,545 fonts; they include 134 font families with light, medium, and bold versions. We specifically use these $134 \mathrm{(families)}\times 3 \mathrm{(weights)}$ fonts for quantitative evaluation with the expectation that the interpolated image of the light and bold versions will be closer to the medium version.

\subsection{Implementation Details}  
\subsubsection{Diffusion Model Architecture}  
We use the architecture of the denoising U-Net in \cite{ho2020denoising} with a slight modification to accept the character class condition $c$ and style feature vector $\mathbf{s}$. The character condition is a 26-dimensional one-hot vector for 
26 capital Latin alphabet classes (`A,' $\ldots$, `Z'). The style feature is a 512-dimensional vector as detailed in Section~\ref{sec:style-feature-extraction}. 
Following \cite{yang2023fontdiffuser}, sampling in the denoising process is performed as
$$
\boldsymbol{\epsilon}_\theta(\mathbf{x}_t \mid t, c, \mathbf{s})=(1-w)\boldsymbol{\epsilon}_\theta(\mathbf{x}_t \mid t)+w\boldsymbol{\epsilon}_\theta(\mathbf{x}_t \mid t, c, \mathbf{s}),
$$
where $w$ is the guidance scale and is set at $3.0$ in the latter experiment. The null tokens are omitted in the first term like \cite{ho2022classifier}.

\subsubsection{Training Diffusion Models}  
U-Net for the diffusion model was trained on the training set of the MyFonts dataset with a batch size of 256 at one million iterations.
We set the sampling steps $T=1,000$ and used the cosine schedule~\cite{nichol2021improved} for the noise schedule.
We used Adam optimizer with a learning rate at $10^{-4}$.
For generating various fonts, we employed data augmentation by horizontal and vertical shift used in~\cite{shirakawa2023ambigram}.
This augmentation is adapted to 30\% of training data, and the shift amount was randomly determined within 20\% of the image size.

\begin{figure}[t] 
\centering
\vspace{3mm}
\includegraphics[width=\textwidth]{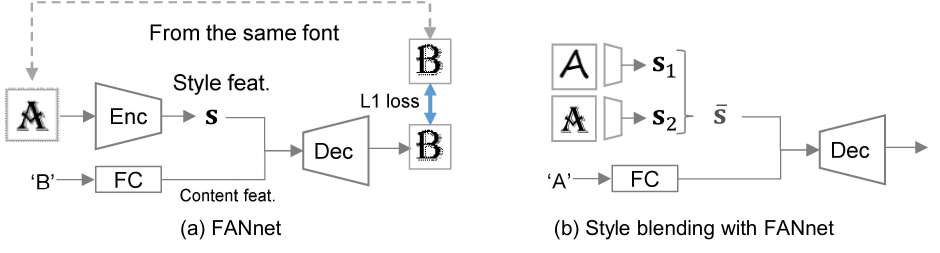}
\caption{(a)~Overview of FANnet~\cite{Roy_2020_CVPR}, which is trained to internally extract the style feature $\mathbf{s}$. (b)~Our comparative model by FANnet. A blended style feature is used to generate an interpolated image.} \label{fig:style-feature-extraction}
\end{figure} 

\subsubsection{Style Feature Extraction\label{sec:style-feature-extraction}}  
We employed FANnet (of STEFANN~\cite{Roy_2020_CVPR}) for extracting a style features $\mathbf{s}$ as a 512-dimensional vector from each character image. Fig.~\ref{fig:style-feature-extraction}~(a) shows the overview of FANnet.
FANnet is a standard network comprised of an encoder and a decoder. They are trained to convert an input character image into a different character image while keeping the input style. For this task, the CNN encoder is expected to extract the input style $\mathbf{s}$. We used CNNs that added batch normalization to the original implementation and trained them with the MyFonts training and validation sets.
\par

\subsubsection{Comparative Model}  
We also employ FANnet as a comparative model because it is still one of the state-of-the-art models for the few-shot font generation of Latin alphabets. As shown in Fig.~\ref{fig:style-feature-extraction}~(b), we utilize the trained FANnet to accept a blended style feature. Specifically, after training the model with the MyFonts training and validation sets, we first extract style features $\mathbf{s}_1$ and $\mathbf{s}_2$ from $\mathbf{r}_1$ and $\mathbf{r}_2$, respectively by the CNN encoder. Then, we blend font style features as $\bar{\mathbf{s}}$ like our condition-blending. Finally, we generate an interpolated image by feeding $\bar{\mathbf{s}}$ to the decoder.

\subsection{Qualitative Evaluation}  
\subsubsection{Interpolating different stroke widths\label{sec:style-parameter}}  
\begin{figure}[t] 
\includegraphics[width=0.95\textwidth]{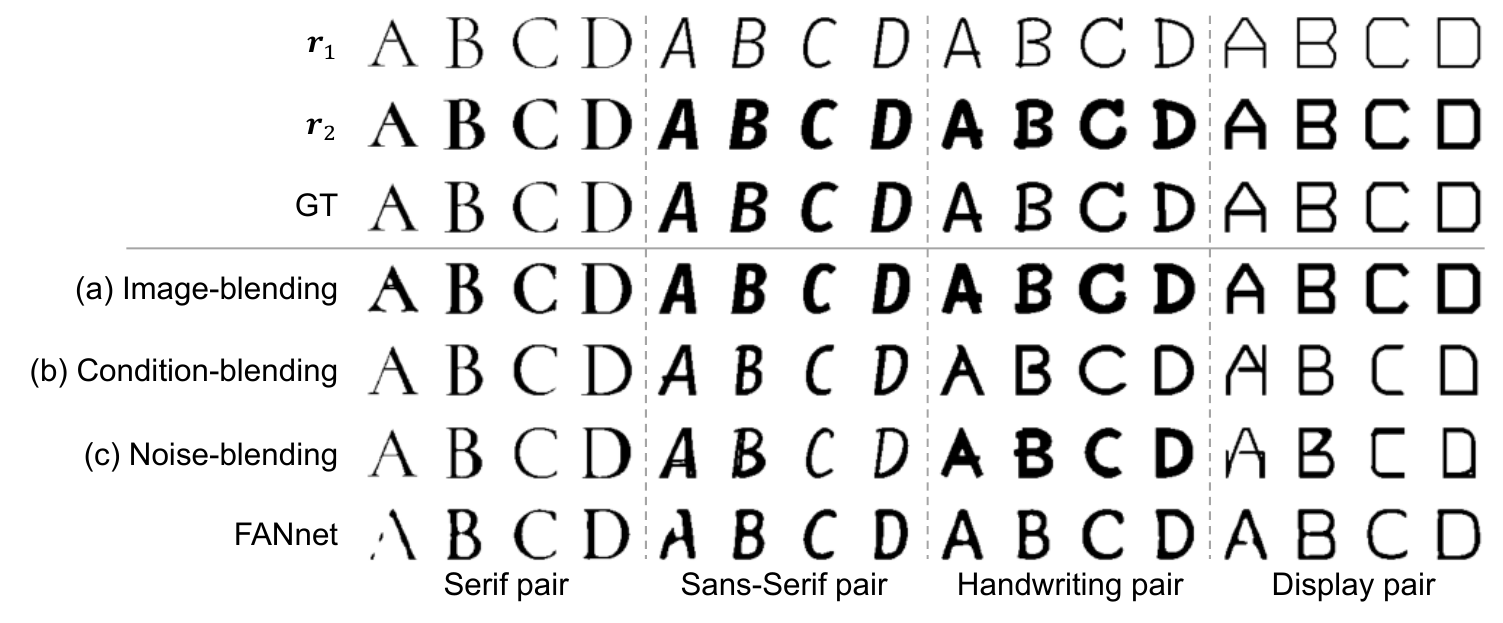}\\[-6mm] 
\caption{Interpolation between the light and bold versions of the same font family in the GoogleFonts dataset. The medium version (GT) is shown as a quasi-ground-truth.\label{fig:qualitative-style-parameter}}
\bigskip
\includegraphics[width=0.95\textwidth]{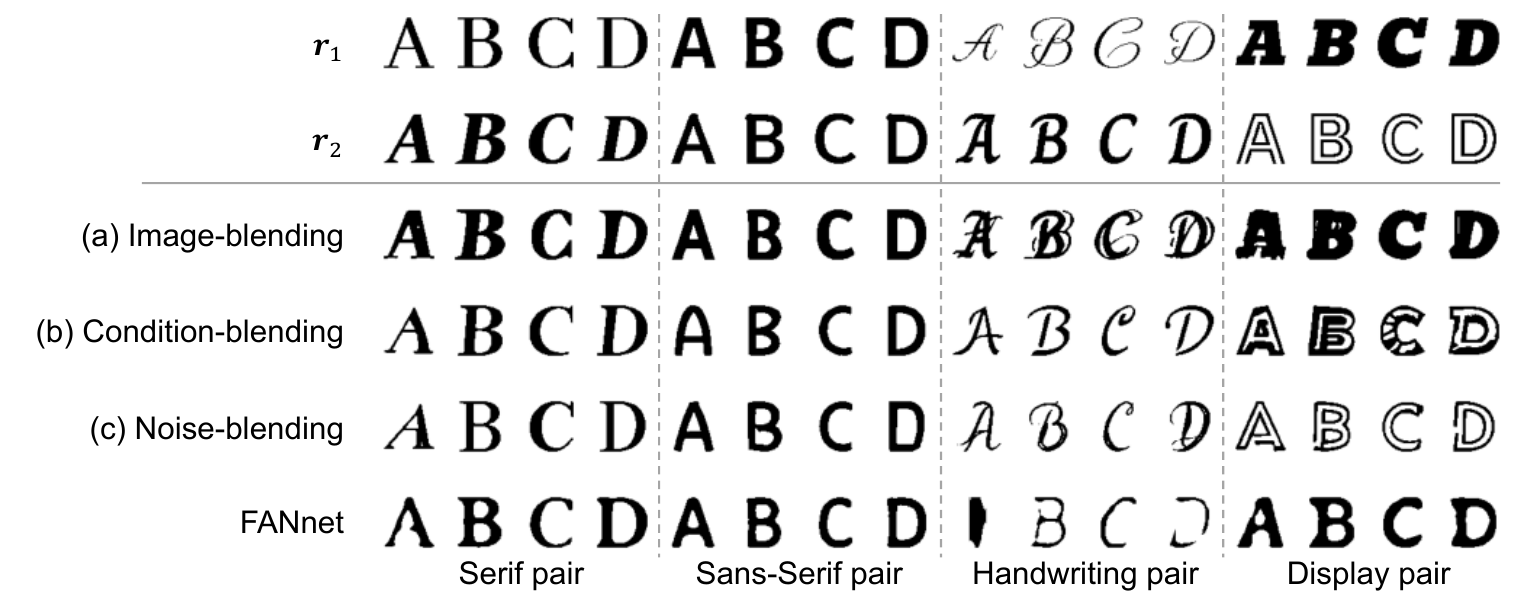}\\[-6mm] 
\caption{Interpolation within the same font style category. (For example, both $\mathbf{r}_1$ and $\mathbf{r}_2$ are sans-serif.)\label{fig:qualitative-similar}}
\end{figure}

Fig.~\ref{fig:qualitative-style-parameter} shows the results of interpolating 
the light version $\mathbf{r}_1$ and the bold version $\mathbf{r}_2$ of the same font family from the GoogleFonts dataset.
The interpolated result with $\lambda=0.5$ is expected to be similar to the medium version, shown as ``GT.''  Since $\mathbf{r}_1$ and $\mathbf{r}_2$ differ only in their stroke widths, it is a rather easy interpolation task.
\par

Figs.~\ref{fig:qualitative-style-parameter}~(a)-(c) show the interpolated results by the three approaches. All interpolated images by the three blending approaches are readable as the original character class. Moreover, the interpolated results keep their original style except for the spurious strokes in Display pairs. In the image-blending approach, $\mathbf{r}_2$ (with thicker strokes) has a larger influence than $\mathbf{r}_1$  (with thinner strokes); this is because the blended image is almost identical to $\mathbf{r}_2$. 
\par

\begin{figure}[t] 
\includegraphics[width=0.95\textwidth]{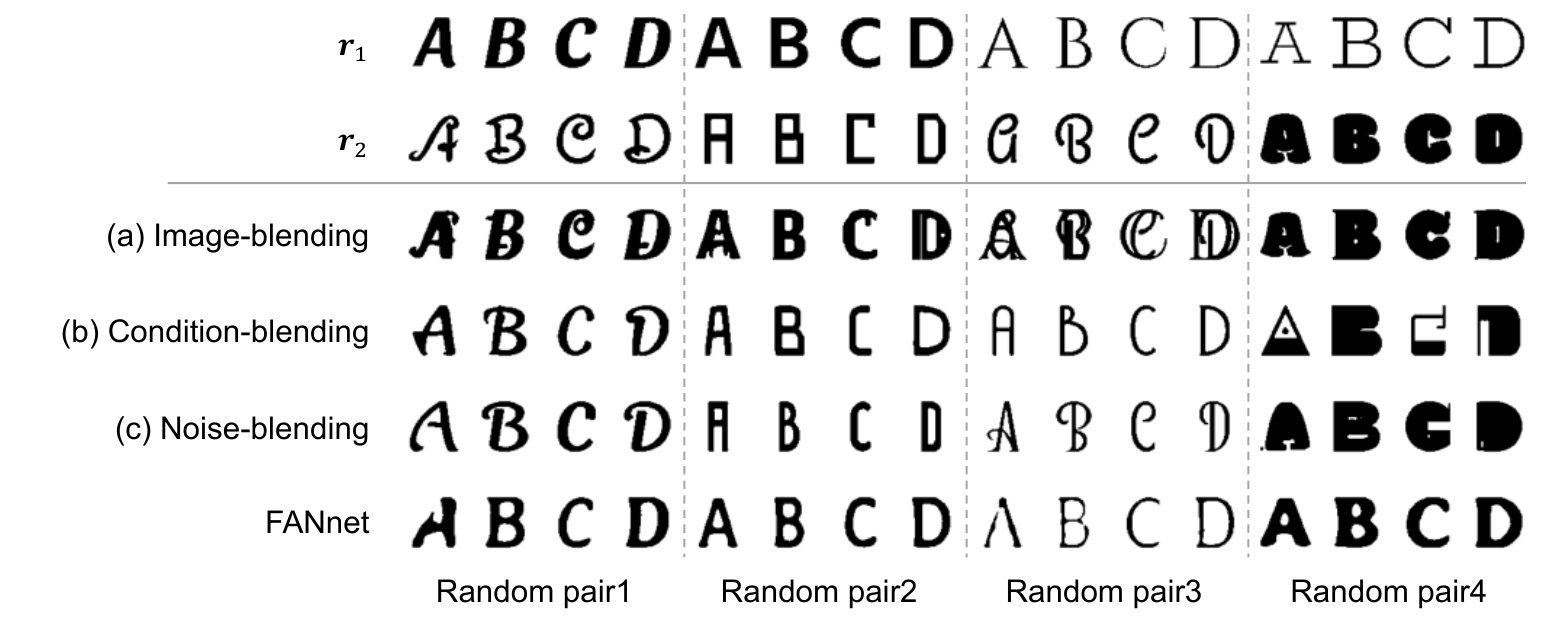} 
\caption{Interpolation between different font style categories.  (For example, $\mathbf{r}_1$ is sans-serif, whereas $\mathbf{r}_2$ is handwriting.)} \label{fig:qualitative-different}
\end{figure} 
\subsubsection{Interpolating within the Same Font Style Category}  
Fig.~\ref{fig:qualitative-similar} shows interpolation results in a more difficult case where $\mathbf{r}_1$ and $\mathbf{r}_2$ come from the same style category but have different styles. Character images from GoogleFonts are used as $\mathbf{r}_1$ and $\mathbf{r}_2$ and interpolated at $\lambda=0.5$. All three approaches preserve the original style category in the interpolation results. Moreover, the interpolation results from the handwriting and display categories often successfully inherit the style characteristics of both reference images.
\par

\subsubsection{Interpolating between Different Font Style Categories}  
Fig.~\ref{fig:qualitative-different} shows interpolation results in a more difficult case where $\mathbf{r}_1$ and $\mathbf{r}_2$ come from the different style categories. Character images from GoogleFonts are randomly selected and used as $\mathbf{r}_1$ and $\mathbf{r}_2$; then they are interpolated at $\lambda=0.5$. 
The results of all three approaches are still readable, and especially the condition-blending approach still preserves both reference styles well, even when 
$\mathbf{r}_1$ and $\mathbf{r}_2$ are very different (``Random pair 4'').
\par

\begin{figure}[t] 
\includegraphics[width=\textwidth]{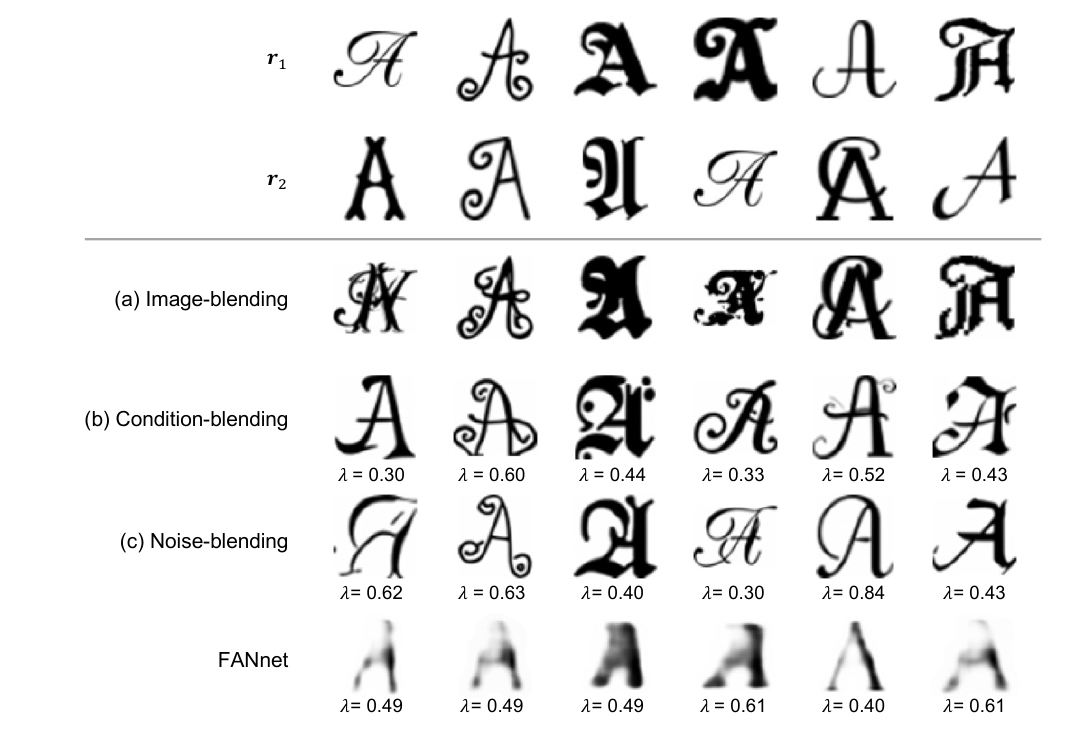} 
\caption{Interpolation of very decorative font styles from MyFonts.} \label{fig:qualitative-decorative}
\end{figure} 
%
\subsubsection{Interpolating Very Decorative Font Styles}  
Fig.~\ref{fig:qualitative-decorative} shows the interpolation results between very decorative font styles from the MyFonts test set. This will be the hardest case for the style interpolation. Surprisingly, all three blending approaches (a)-(c) could manage this hard case, and the interpolated results still show a reasonable mixture of the reference styles. Moreover, except for a few cases, the readability of the letter `A' is also preserved. Since different results are given by different approaches, it is meaningful to examine all approaches when we want to find brand-new styles by interpolating existing fonts.\par 

\subsubsection{Effect of $\lambda$}  
\begin{figure}[t] 
\centering
\includegraphics[width=0.95\textwidth]{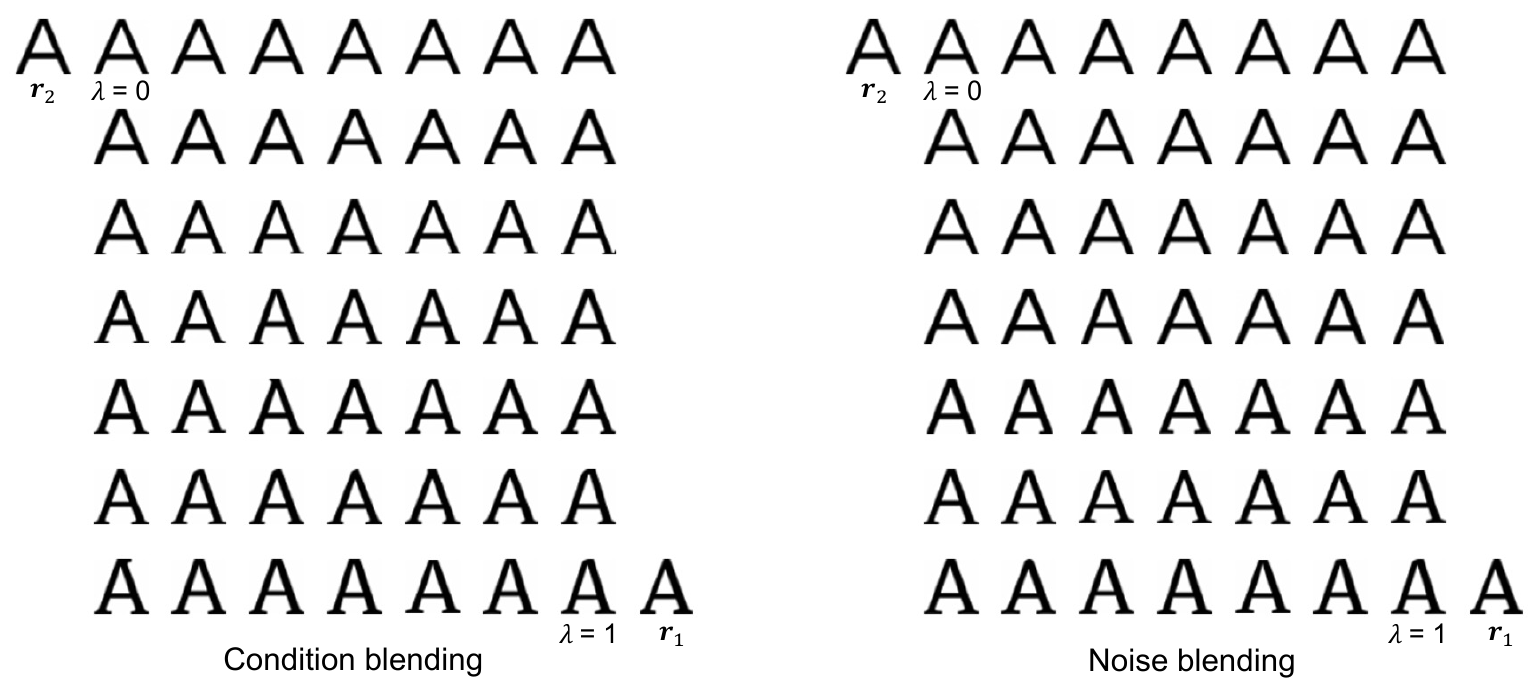} 
\includegraphics[width=0.95\textwidth]{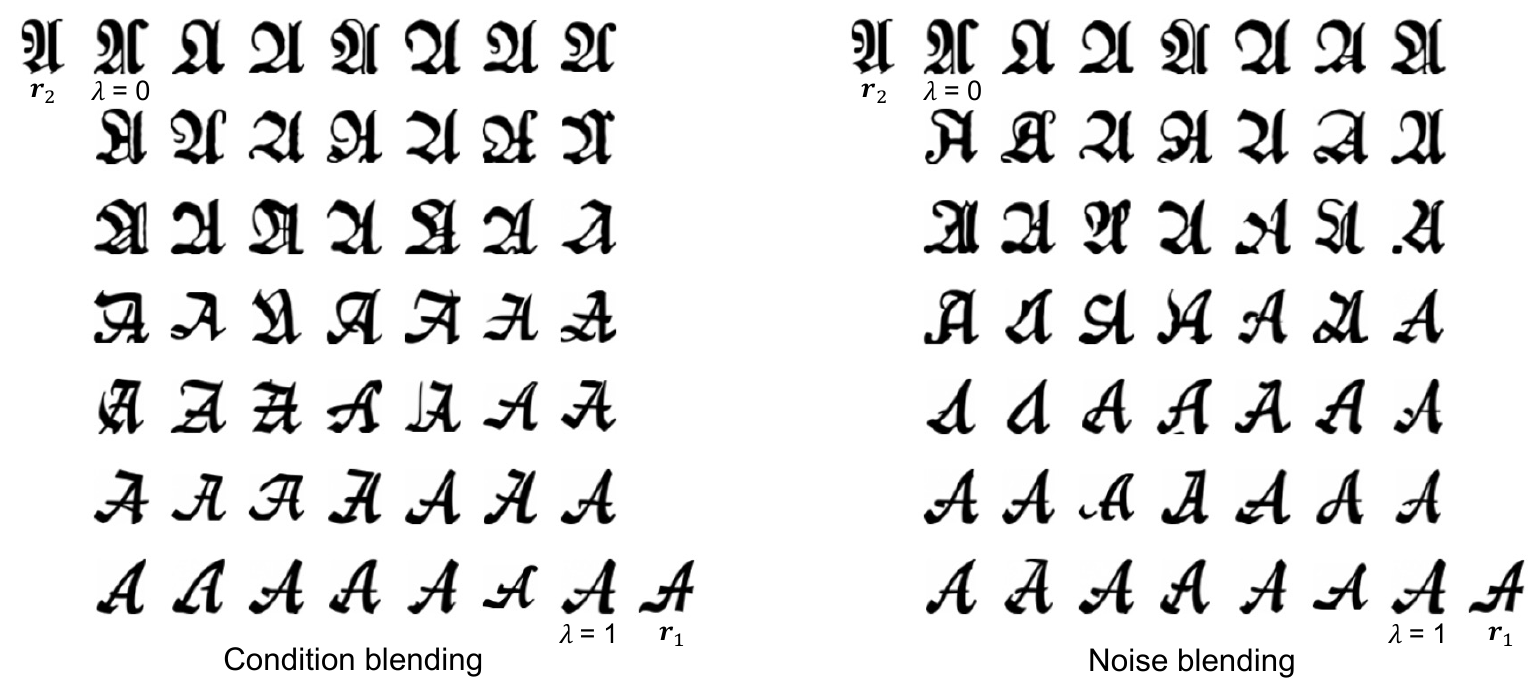}\\[-3mm] 
\caption{Interpolation with different values of $\lambda$. From the top-left to the bottom-right, the value of $\lambda$ changes from 0 to 1 with the same interval (of $1/48$). 
\label{fig:qualitative-various-lambda}}
\end{figure} 

Fig.~\ref{fig:qualitative-various-lambda} shows the interpolated images generated while changing $\lambda$ from 0 to 1 with the same interval by the condition-blending and noise-blending approaches. 
Here, the condition-blending and noise-blending approaches are examined on two pairs of `A' from 
the MyFonts test set. In the upper example with $\mathbf{r}_1$ (sans-serif) and $\mathbf{r}_2$ (serif), both blending approaches show a near-linear behavior about serif; the interpolated results show a good mixture of two styles with a short serif. In contrast, the horizontal stroke of `A' sometimes shows changes in its height. This fact suggests that the style condition space and the noise space are not very linear to the resulting letter shapes.\par  

In the lower example with decorative references, $\mathbf{r}_1$ (black-letter) and $\mathbf{r}_2$ (calligraphic), both approaches show more nonlinear behaviors along with $\lambda$. In fact, the interpolated images show an abrupt change in their styles. At several $\lambda$ values, they generate serendipitous styles. From this result, we also can say that the style interpolation with diffusion models will inspire new font designs.\par 

\subsubsection{Qualitative Comparison with FANnet}  
In the above interpolation results, we also show the results given by FANnet.
When interpolating standard styles (such as Fig.~\ref{fig:qualitative-style-parameter} and the serif and sans-serif styles of Fig.~\ref{fig:qualitative-similar}), FANnet also shows reasonable interpolation results without large degradation. However, when interpolating decorative styles
or very different styles, such as Fig.~\ref{fig:qualitative-decorative}, 
FANnet generates blurred images whose styles are irrelevant to the reference 
images. A more careful observation suggests that FANnet can only generate similar standard styles. Although it is a good strategy to preserve readability, it is inappropriate for finding new designs by interpolation. Later quantitative evaluations will also confirm this conservative behavior of FANnet.
\par

\begin{table}[t]
\caption{Recognition accuracy $\uparrow$ (\%) of 13,000 interpolated character images generated from the MyFonts test set. The distribution-level recall and precision\cite{kynkaanniemi2019improved} are also evaluated. The \red{very low recall} by FANnet suggests that FANnet only provides similar interpolated images regardless of the input pairs.}
\label{table:recognition-accuracy-myfonts} 
\centering
\begin{tabular}{l||r|r|r} \hline
  Approach    & Accuracy $\uparrow$ & Precision  $\uparrow$ & Recall  $\uparrow$ \\ \hline
  FANnet\cite{Roy_2020_CVPR} &  85.7  &  \textbf{0.660}     &   \red{0.0506}  \\
  Image-blending &  85.1  &   0.480  &   0.649 \\
  Condition-blending &  \textbf{87.7} &   0.539    & 0.676 \\ 
  Noise-blending &  85.6  &    0.548   &   \textbf{0.709} \\ \hline
\end{tabular}\\[5mm]
\end{table} 

\begin{table}[t]
\caption{Recognition accuracy $\uparrow$  (\%) of character images generated. See the main text for the details.}
\label{table:recognition-accuracy} 
\centering
\begin{tabular}{l||>{\raggedleft}p{1cm}|>{\raggedleft}p{1cm}|>{\raggedleft}p{1cm}|>{\raggedleft}p{1cm}|>{\raggedleft}p{1cm}|r} \hline
            & \multicolumn{5}{c|}{Interpolated  pairs}&\\ \cline{2-6} 
  Approach    & S-S  & SS-SS & H-H & D-D & R-R & Average \\ \hline
  Image-blending &  \textbf{98.9}  &   \textbf{99.4}  &   84.7   &   91.7   &  96.4  &  94.2 \\
  Condition-blending &  98.5 &   98.3    & \textbf{89.5} &   \textbf{93.3}   &  \textbf{96.8} &  \textbf{95.3} \\ 
  Noise-blending &  97.4  &    97.1   &   86.7   &   93.2   &  95.0  &  93.9 \\  \hline
\end{tabular}
\end{table} 
\subsection{Quantitative Evaluation: Generated Character Recognition}  
To quantify the readability of the generated characters, we evaluated them in a character recognition task. When two font images of the same character class are interpolated appropriately, the generated font image is expected to have the same character class. We, therefore, conducted 26-class (`A'--`Z' ) character recognition by ResNet-18~\cite{he2016deep}. If the recognition accuracy is high, the generated images will have better readability, i.e., better quality. ResNet-18 was trained by the training and validation sets from the MyFonts dataset. The model achieved 91.6\% accuracy on original character images in the MyFonts test set.\par

Table~\ref{table:recognition-accuracy-myfonts} shows the character recognition accuracy on the 13,000 interpolated images at $\lambda=0.5$. We randomly selected 500 character pairs for each of the 26 classes from the MyFonts test set and then generated the interpolated image for each pair. The fonts of paired characters are random; therefore, the fonts of two paired `A's can be very different. All three blending approaches and FANnet achieved almost the same recognition accuracies --- around 85.1\% (for image-blending) and 87.7\% (for condition-blending). Since the accuracy of the original (i.e., non-interpolated) images is 91.6\%, these accuracies indicate that all approaches generate fairly readable character images. Note that some generated images are very decorative by interpolating pairs of different decorative fonts like Fig.~\ref{fig:qualitative-decorative}.
\par 

The comparative model, FANnet, performs poorly, although it gives a similar accuracy to our approaches. To show this fact, we evaluate the distribution-level recall and precision\cite{kynkaanniemi2019improved}. These metrics are useful to understand how the distribution of the interpolated images is similar to that of the original images. The recall $\in [0,1]$ measures how the distribution of the interpolated images covers that of the original images, and the precision  $\in [0,1]$ measures how the latter distribution covers the former. The very low recall by FANnet shows that the interpolated images by FANnet are not diverse enough and result in similar (near-standard) character shapes. This behavior increases recognition accuracy but is totally undesirable for interpolation-based novel character image generation. In contrast, our approaches achieve far higher recall values; namely, the characters by our approaches have enough diversity while preserving high readability. 
\par

Table~\ref{table:recognition-accuracy} shows the recognition accuracy of the interpolation characters with the GoogleFonts dataset at $\lambda=0.5$. Here, S, SS, H, and D denote style categories: Serif, Sans-Serif, Handwriting, and Display (Decorative). ``S-S'' means Serif pairs. For each font category and each character class, we prepared 2,600 interpolated images. ``R-R'' means 2,600 randomly chosen pairs (per class) regardless of font categories. This category-wise evaluation also reveals all approaches generate character images with high readability. It is found that interpolation of ``H-H'' is even harder than ``D-D.'' A closer inspection reveals that for difficult pairs, such as  ``H-H'' and ``D-D,'' the image-blending approach is inferior to the others.

\subsection{Quantitative Evaluation: Interpolating Different Stroke Widths}
Finally, we conduct a finer shape evaluation of the interpolated images by using the font family with different stroke widths. As noted in Section~\ref{sec:datasets}, we have 134 font families with light, medium, and bold versions in the GoogleFonts dataset. As shown in Fig.~\ref{fig:qualitative-style-parameter}, the interpolation results of the light and bold versions (at $\lambda=0.5$) are expected to be similar to the medium version. Based on this expectation, we assume the medium version is a quasi-ground-truth of the interpolation result and compare it with the interpolated result by the pixel-wise evaluation with MSE. Since 134 families comprise 28 serif, 99 sans-serif, 2 handwriting, and 5 display fonts, the MSE is evaluated on 728, 2,574, 52, and 130 interpolated images.\par

Table~\ref{table:average-similarity} shows the results. As expected from Fig.~\ref{fig:qualitative-style-parameter}, the image-blending approach gives the best performance; however, it is not a meaningful result because the blended image is almost identical to the bold version and the MSE evaluates the difference between the medium and bold versions of the same font. A more important result is the comparison between condition-blending and noise-blending; the former is slightly better than the latter in all font style categories.\par
\begin{table}[t]
\caption{Average MSE $\downarrow$ using the character images with three different stroke widths, light, medium, and bold. Note that better performance by image-blending is not meaningful in this evaluation scheme. (See the main text for details.) }
\label{table:average-similarity} 
\centering
\begin{tabular}{l||>{\raggedleft}p{1cm}|>{\raggedleft}p{1cm}|>{\raggedleft}p{1cm}|>{\raggedleft}p{1cm}|r} \hline
            & \multicolumn{4}{c|}{Interpolated pairs}&\\   \cline{2-6}
  Approach    & S-S  & SS-SS & H-H & D-D &  Average \\ \hline
  Image blending & 0.106 &  0.116   & 0.123  & 0.150   & 0.115   \\
  Condition blending &  0.125  &   0.128    &   0.161   &   0.167   &   0.129   \\
  Noise blending &  0.136  &   0.138    &   0.179   &   0.175    &   0.139  \\ \hline
\end{tabular}
\end{table} 
\subsection{Summary of Experimental Evaluations\label{sec:ex-summary}}
Overall, the qualitative comparisons between the three blending approaches show that condition-blending and noise-blending show their great ability to generate interpolated images. Given a pair of largely different shapes, they can even generate serendipitous results. The image-blending approach is not bad but has a limitation: fonts with thicker strokes dominate those with lighter strokes.
The quantitative comparisons also did not show any large difference in their performance. However, a precise observation reveals that the condition-blending approach generates the most readable images with the highest recognition accuracy. At the same time, since the recall of the condition-blending approach is slightly lower than the noise-blending, the former seems slightly more conservative than the latter.

\section{Conclusion, Limitation, and Future Work}
This paper uses diffusion models to generate new font styles by interpolating two reference character images in the same character class but with different font styles. Thanks to the flexibility of diffusion models, we could consider three interpolation approaches: image-blending, condition-blending, and noise-blending approaches. We could confirm that the latter two approaches generate readable interpolated images even with very different styles through various qualitative and quantitative evaluations. The serendipitous interpolated images show that our interpolation approaches will provide various hints for brand-new font designs. \par

The current limitation is about the setting of the interpolation parameter $\lambda$. We observed that there sometimes happens a ``jump'' between the interpolated images with slightly different $\lambda$ values. If a more stable generation without jumps is necessary, the  denoising network model
(i.e., U-Net $\theta$ in Fig.~\ref{fig:overview}~Denoising process) needs to be retrained with special loss functions so that its latent space becomes smoother. Another future work candidate is a set-wise interpolation of all alphabets (i.e., \{`A'-`Z'\} $\leftrightarrow$ \{`A'-`Z'\}) instead of the current letter-wise interpolation (i.e., `A'$\leftrightarrow$ `A', $\ldots$, `Z'$\leftrightarrow$`Z'). By the set-wise interpolation, we can expect style-consistent interpolation results for all alphabet letters. Our current target is limited to font images; however, our interpolation approaches are applicable to arbitrary images, such as standard object images, medical images, and artistic images, which can be our future targets.
\par
\bigskip
\noindent{\bf Acknowledgment}:\ This work was supported by JSPS KAKENHI Grant Number JP22H00540.

\bibliographystyle{splncs04}
\bibliography{ref}

\end{document}